\documentclass[runningheads,a4paper]{llncs}

    \usepackage{amssymb}
    \usepackage{amsmath}
    \usepackage[misc,geometry]{ifsym}
    \usepackage{graphicx}
    \usepackage{caption}
    \graphicspath{{images/}}

    \usepackage{subfigure}
    \usepackage{booktabs}
    
    \usepackage{url}
    \urldef{\mailsc}\path|haofu.liao@rochester.com|

\begin{document}
    
    \mainmatter  
    
    \title{More Knowledge is Better: Cross-modality Volume Completion and 3D+2D Segmentation for Intracardiac Echocardiography Contouring}
    
    \titlerunning{Intracardiac echocardiography contouring}
    
    %
    %
    \author{Haofu Liao\textsuperscript{1,2(\Letter)}
    \and Yucheng Tang\textsuperscript{2,3}
    \and Gareth Funka-Lea\textsuperscript{2}
    \and Jiebo Luo\textsuperscript{1}
    \and \\ Shaohua Kevin Zhou\textsuperscript{2}}
    \authorrunning{H. Liao et al}
   
    
    \institute{\textsuperscript{1} Department of Computer Science, University of Rochester, Rochester, USA\\ \mailsc\\ 
    \textsuperscript{2} Medical Imaging Technologies,
    Siemens Healthineers, Princeton, USA \\
    \textsuperscript{3} Department of ECE,
    New York University, New York, USA}
   
    %
    %
    
    \toctitle{}
    \tocauthor{H. Liao et al}
    \maketitle

    \begin{abstract}
    Using catheter ablation to treat atrial fibrillation increasingly relies on intracardiac echocardiography (ICE) for an anatomical delineation of the left atrium and the pulmonary veins that enter the atrium. However, it is a challenge to build an automatic contouring algorithm 
    because ICE is noisy and provides only a limited 2D view of the 3D anatomy. This work provides \textit{the first automatic solution} to segment the left atrium and the pulmonary veins from ICE. In this solution, we demonstrate the benefit of building a \textit{cross-modality framework} that can leverage a database of diagnostic images to supplement the less available interventional images. To this end, we develop a novel deep neural network approach that uses the (i) \textit{3D geometrical information} provided by a position sensor embedded in the ICE catheter and the (ii) \textit{3D image appearance information} from a set of computed tomography cardiac volumes. 
    We evaluate the proposed approach over 11,000 ICE images collected from 150 clinical patients. Experimental results show that our model is significantly better than a direct 2D image-to-image deep neural network segmentation, especially for less-observed structures.
    \end{abstract}
    
\section{Introduction}

Atrial fibrillation (AF)
affects about 2\% to 3\% of the population in Europe and North America as of 2014  \cite{zoni2014}. One of its treatments is to perform catheter ablation to destroy the atypical tissues.
During catheter ablation, intracardiac echocardiography (ICE) is often used to guide the intervention. Compared with other imaging modalities such as transoesophageal echocardiography, ICE provides better patient tolerance, requiring no general anesthesia \cite{bartel2013intracardiac}. Moreover, modern ICE devices are equipped with an embedded position sensor that measures the precise 3D location of the ICE transducer. Such spatial geometry information associated with the ICE image is key to this study.

\begin{figure}[t]
    \centering
    \subfigure[]{
        \includegraphics[width=0.2475\textwidth]{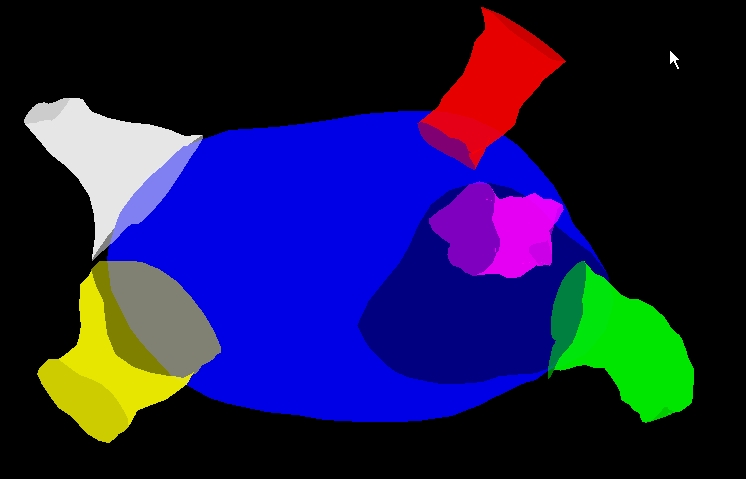}
    }
    \hspace*{1em}
    \subfigure[]{
        \includegraphics[width=0.5625\textwidth]{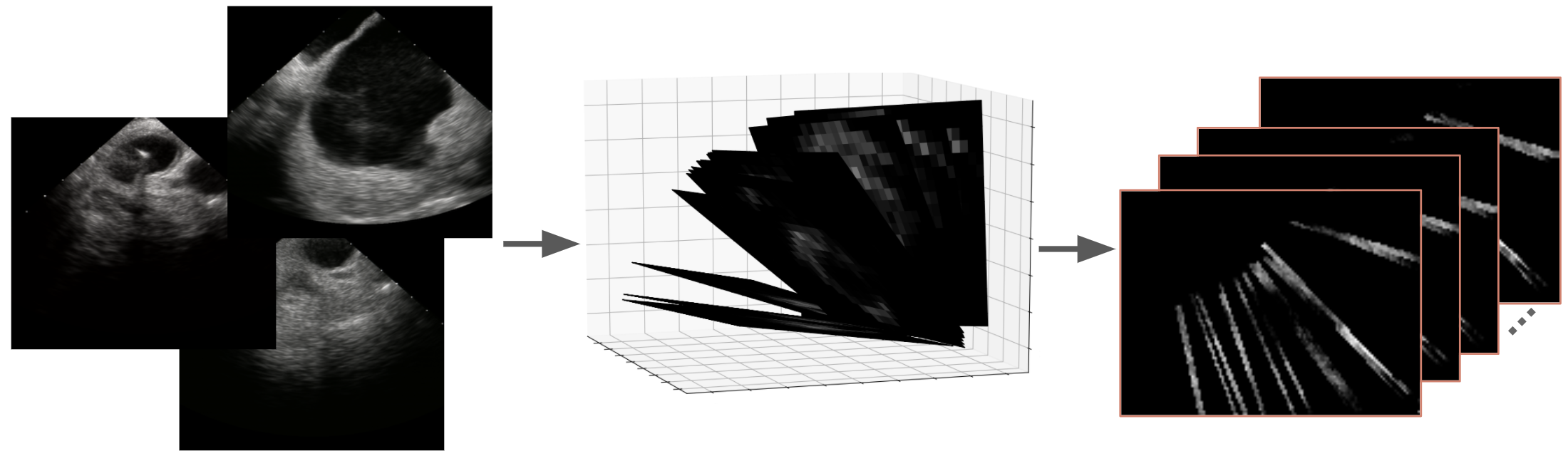}
    }
    \caption{(a) Graphical illustration of LA and its surrounding structures: blue-LA, green-left atrial appendage (LAA), red-left inferior pulmonary vein (LIPV), purple-left superior pulmonary vein (LSPV), white-right inferior pulmonary vein (RIPV), yellow-right superior pulmonary vein (RSPV). (b) 3D sparse ICE volume generation using the location information associated with each ICE image.}
    \label{fig:sparse_usd}
\end{figure}

Some gross morphological and architectural features of the left atrium (LA)
are important to AF interventions and recognizing these features relies on a clear view of LA's surrounding structures (see Fig. \ref{fig:sparse_usd} (a)) and their junctions with the LA \cite{sanchez2014left}. However, due to the limitations of 2D ICE and the difficulty in manual manipulation of the ICE transducer, these 3D anatomical structures may not be sufficiently observed in certain views. This introduces difficulties to electrophysiologists as well as echocardiography image analysis algorithms that attempt automatic multi-component contouring or segmentation.


Existing approaches to 2D echocardiogram segmentation only focus on single cardiac chamber such as left ventricle (LV) \cite{lin2003combinative,sarti2005maximum,zhou2010shape} or LA \cite{allan2017simultaneous}. They are designed to distinguish between the blood tissues and the endocardial structures which is relatively easy due to the significant difference in appearance. When it comes to multiple cardiac components (chambers and their surrounding structures), where the boundaries cannot be clearly recognized, these methods may fail. To the best of our knowledge, this paper is the first to handle the multi-component echocardiogram segmentation from 2D ICE images.


Recently, deep convolutional neural networks (CNNs) have achieved unprecedented success in medical image analysis, including segmentation \cite{zhou17ap}. 
However, our baseline method of training a CNN to directly generate segmentation masks from 2D ICE images does not demonstrate satisfactory performance, especially for the less-observed pulmonary veins . Such a baseline solely relies on the brute force of big data to cover all possible variations, which is difficult to achieve. To go beyond brute force, we further integrate knowledge to boost contouring performance. \footnote{The outcome of this research has been patented \cite{funka2019three}.}  Such knowledge stems from two sources: (i) \textit{3D geometry information} provided by a position sensor embedded inside an ICE catheter, and (ii) \textit{3D image appearance information} exemplified by cross-modality computed tomography (CT) volumes that contain the same anatomical structures.

\section{Method}

The proposed method consists of three parts. Using the 3D geometry knowledge, we first form a 3D sparse volume based on the 2D ICE images. Then, to tap into the 3D image appearance knowledge, we design a multi-task 3D network with an adversarial formulation. The network performs cross-modality volume completion and sparse volume segmentation simultaneously for collaborative structural understanding and consistency. Finally, taking as inputs both the original 2D ICE image and the 2D mask projected from the generated 3D mask, we design a network to refine the 2D segmentation results.




We form a 3D sparse ICE volume from a set of 2D ICE images with each including part of the heart in its field of view and its 3D position from a magnetic localization system. As shown in Fig. \ref{fig:sparse_usd}, we use the location information to map all ICE images (left) to 3D space (middle), thus forming a sparse ICE volume (right). The generated sparse ICE volume keeps the spatial relationships among individual ICE views. A segmentation method based on the sparse volume can take this advantage for better anatomical understanding and consistency.

\subsection{3D Sparse Volume Segmentation and Completion}

\begin{figure}[t]
        \centering
        \subfigure[3D-SCNet]{
        \includegraphics[width=0.5\textwidth]{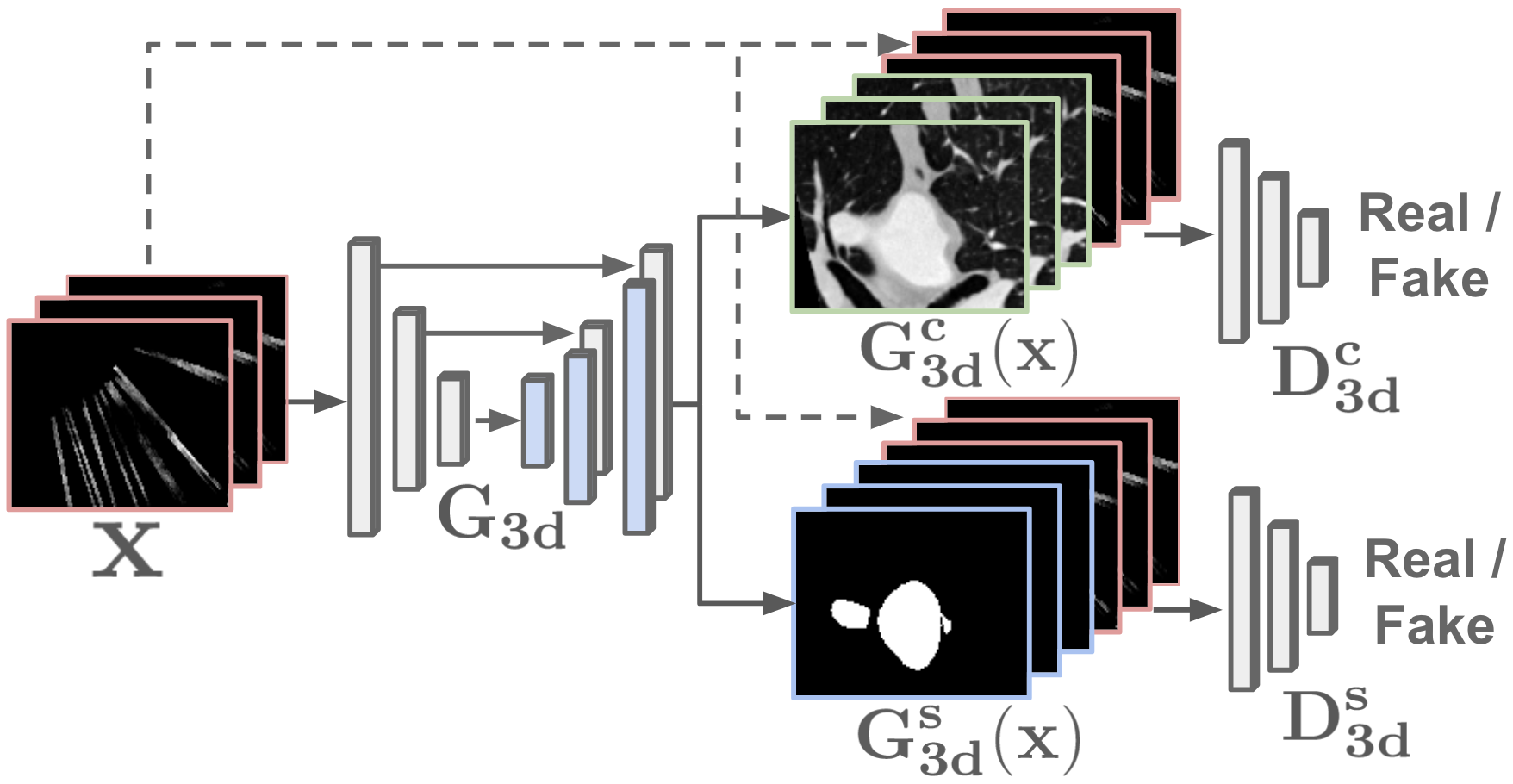}
        }
        \subfigure[2D-RefineNet]{
        \includegraphics[width=0.45\textwidth]{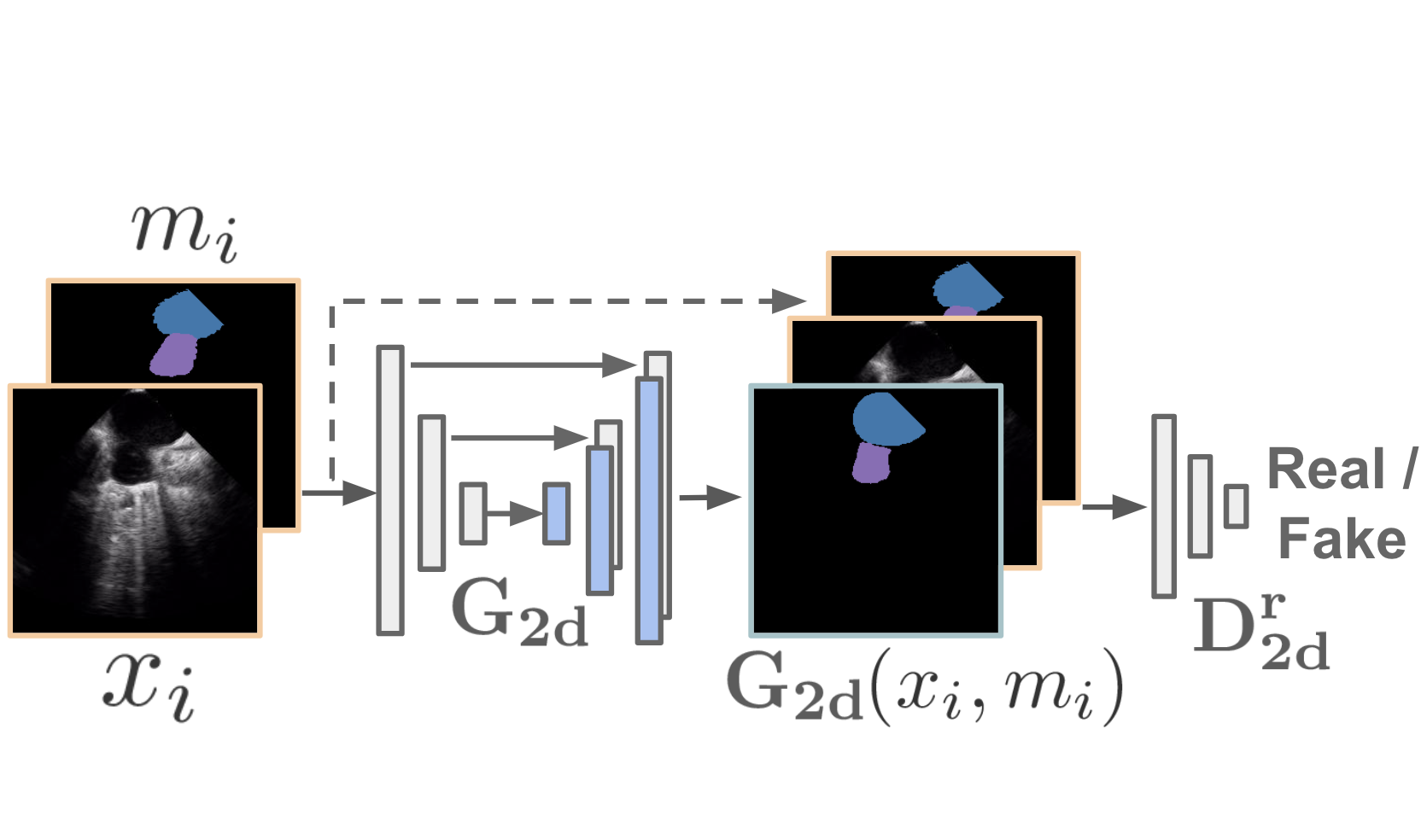}
        }
        \caption{The network architectures of the proposed method.}
        \label{fig:architecture}
    \end{figure}

The architecture of the proposed 3D segmentation and completion network (3D-SCNet) is illustrated in Fig. \ref{fig:architecture}(a). The network consists of a generator $G_{3d}$ and two discriminators $D_{3d}^c$ and $D_{3d}^s$. Taking the sparse ICE volume $\mathbf{x}$ as input, $G_{3d}$ performs 3D segmentation and completion simultaneously, and outputs a segmentation map $G_{3d}^s(\mathbf{x})$ as well as a dense volume $G_{3d}^c(\mathbf{x})$. During training, the ground truth of $G_{3d}^c(\mathbf{x})$ is a CT volume instead of a dense ICE volume as we lack the training data of the latter. The ICE images and the CT volumes are from \textit{completely different} patients. This inherently indicates a challenging \textit{cross-modality volume completion} problem with unpaired data. We target this problem through adversarial learning and mesh pairing (See Sec. \ref{sec:experiments}). The two discriminators judge the realness of the outputs from the generator. When trained adversarially together with a generator, they make sure the generator's outputs are more perceptually realistic. Following conditional GAN \cite{isola2017image}, we also allow the discriminators to take $\mathbf{x}$ as the input to further improve adversarial training. \vspace{0.5em}

\noindent \textbf{Adversarial loss} The segmentation task $s$ and completion task $c$ are trained jointly in a multi-task learning (MTL) fashion \cite{caruana1998multitask,liao2018face}. The adversarial loss for a task $t \in \{s,c\}$ can be written as
\begin{equation} 
        \label{eq:adversarial}
        \begin{split}
        &\mathcal{L}^{t}_{adv} = \mathbb{E}_{\mathbf{x},y_{t} \sim p (\mathbf{x}, y_{t})}[\log D_{3d}^t(\mathbf{x}, y_t )] + \mathbb{E}_{\mathbf{x} \sim p (\mathbf{x})}[1 - \log D_{3d}^t(\mathbf{x}, G_{3d}^t(\mathbf{x}))],
        \end{split}
\end{equation}
where $p$ denote the data distributions. For a real data $y_t$, i.e., the ground truth segmentation map or CT volume, $D_{3d}^t$ is trained to predict a ``real'' label. For the generated data $G_{3d}^t(\mathbf{x})$, $D_{3d}^t$ learns to give a ``fake'' label. On the other hand, the generator $G_{3d}$ is trained to deceive $D_{3d}^t$ by making $G_{3d}^t(\mathbf{x})$ as ``real'' as possible. \vspace{0.5em}
 
\noindent \textbf{Reconstruction loss} Adversarial loss alone, however, does not give a strong structural regularization to the training \cite{pathak2016context}. Hence, we use reconstruction loss to measure the pixel-level error between the generator outputs and the ground truths. For the segmentation task, we first convert the score map to a multi-channel map with each channel denoting the binary segmentation map of a target anatomy and then apply an L2 loss $\mathcal{L}_{rec}^s$ between $G_{3d}^s(\mathbf{x})$ and $y_s$. For the completion task, the L1 loss $\mathcal{L}_{rec}^c$ between $G_{3d}^c(\mathbf{x})$ and $y_c$ is measured. We use L1 loss against L2 loss for this task due to the observation that outputs from L2 losses are usually overly smoothed. The total loss of the sparse volume segmentation and completion network is given by
\begin{equation}
        \mathcal{L}_{3d} = \sum_{t \in \{c, s\}}\lambda^{t}_{rec} \mathcal{L}^{t}_{rec} + \lambda^{t}_{adv} \mathcal{L}^{t}_{adv},
\end{equation}
where $\lambda_{rec}^t$ and $\lambda_{adv}^t$ balance the importance of the reconstruction loss and reconstruction loss, respectively. \vspace{0.5em}
    
\noindent \textbf{Architecture details} We use a 3D UNet-like network \cite{ronneberger2015u} as the generator. There are 8 consecutive downsampling blocks followed by 8 consecutive upsampling blocks in the network. We use skip connections to shuttle feature maps between two symmetric blocks. Each downsampling block contains a 3D convolutional layer, a batch normalization layer and a leaky ReLU layer. Similarly, each upsampling layer contains a 3D deconvolutional layer, a batch normalization layer and a ReLU layer. The convolutional and deconvolutional layers have the same parameter settings: $4 \times 4 \times 4$ kernel size, $2 \times 2 \times 2$ stride size and $1 \times 1 \times 1$ padding size. Finally, a $\tanh$ function is attached at the end of the generator to bound the network outputs. The two discriminators $D_{3d}^s$ and $D_{3d}^c$ have identical network architecture with each of them having 3 downsampling blocks followed by a 3D convolutional layer and a sigmoid layer. The downsampling blocks for the discriminators are the same as the ones used in the generator. The final 3D convolutional layer ($3 \times 3 \times 3$ kernel size, $1 \times 1 \times 1$ stride size and $1 \times 1 \times 1$ padding size) and sigmoid layer are used for realness classification.

\subsection{2D Contour Refinement}

As shown in Fig. \ref{fig:architecture}(b), the 2D refinement network (2D-RefineNet) has a similar structure to the 3D-SCNet. Actually, $G_{2d}$ and $D_{2d}^r$ have almost the same structure as their 3D counterparts except that the convolutional and deconvolutional layers are now in 2D. The inputs to the 2D-RefineNet is a 2D ICE image $x_i$ together with its corresponding 2D segmentation map $m_i$, where $m_i$ is obtained by projecting $G_{3d}^s(\mathbf{x})$ onto $x_i$. The training of the 2D-RefineNet is also performed in an adversarial fashion and conditional GAN is used to allow $D_{2d}^r$ observing the generator inputs. We compute the adversarial loss $L_{adv}^r$ the same way as Eq. (\ref{eq:adversarial}) and use the L2 distance between the refinement network output $G_{2d}(x_i, m_i)$ and the ground truth 2D segmentation map $y_r$ as the reconstruction loss $L_{rec}^r$. The total loss is
    \begin{equation}
        \mathcal{L}_{2d} =\lambda^{r}_{rec} \mathcal{L}^{r}_{rec} + \lambda^{r}_{adv} \mathcal{L}^{r}_{adv},
    \end{equation}
    where $\lambda^{r}_{rec}$ and $\lambda^{r}_{adv}$ are the corresponding balancing coefficients.

    \section{Experiments} \label{sec:experiments}

    \begin{figure}[t]
        \centering
        \subfigure[]{
        \includegraphics[width=.17\textwidth]{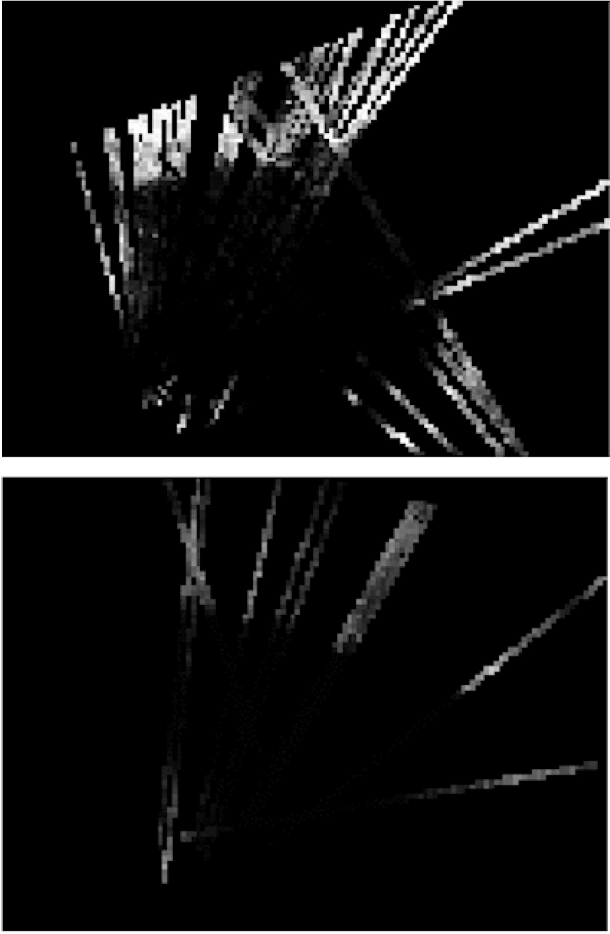}
        }
        \hspace*{-1.1em}
        \subfigure[]{
        \includegraphics[width=.17\textwidth]{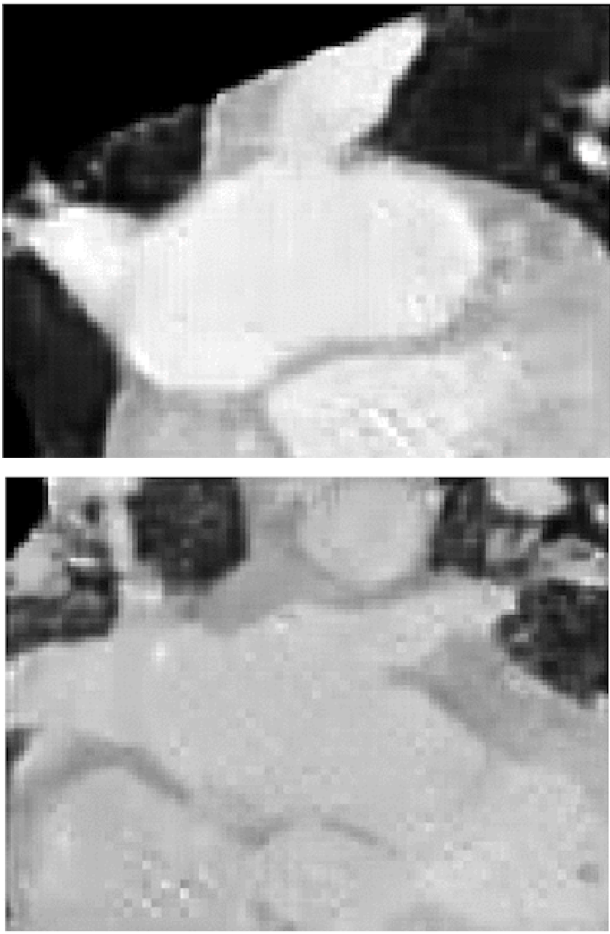}
        }
        \hspace*{-1.1em}
        \subfigure[]{
        \includegraphics[width=.17\textwidth]{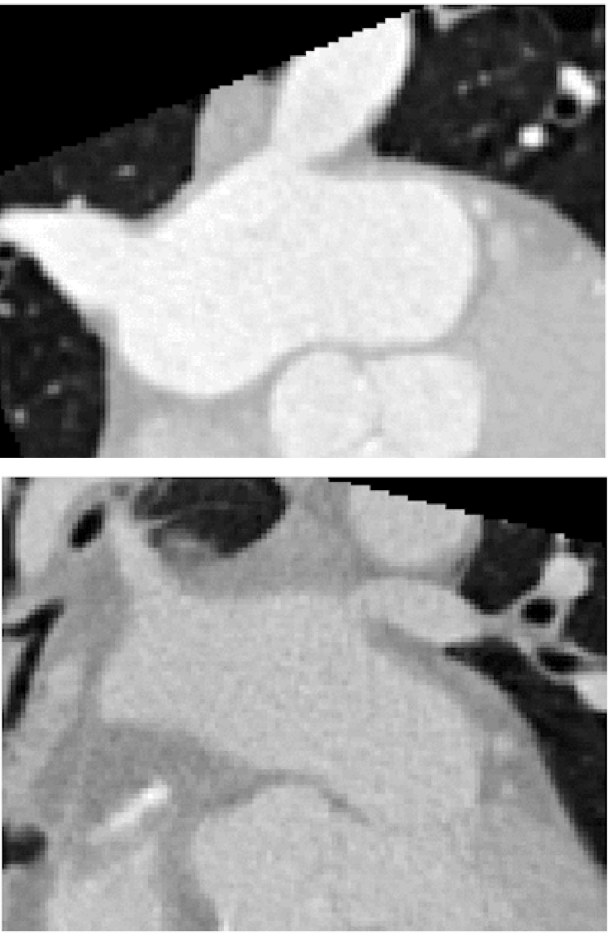}
        }
        \hspace*{-1.1em}
        \subfigure[]{
        \includegraphics[width=.17\textwidth]{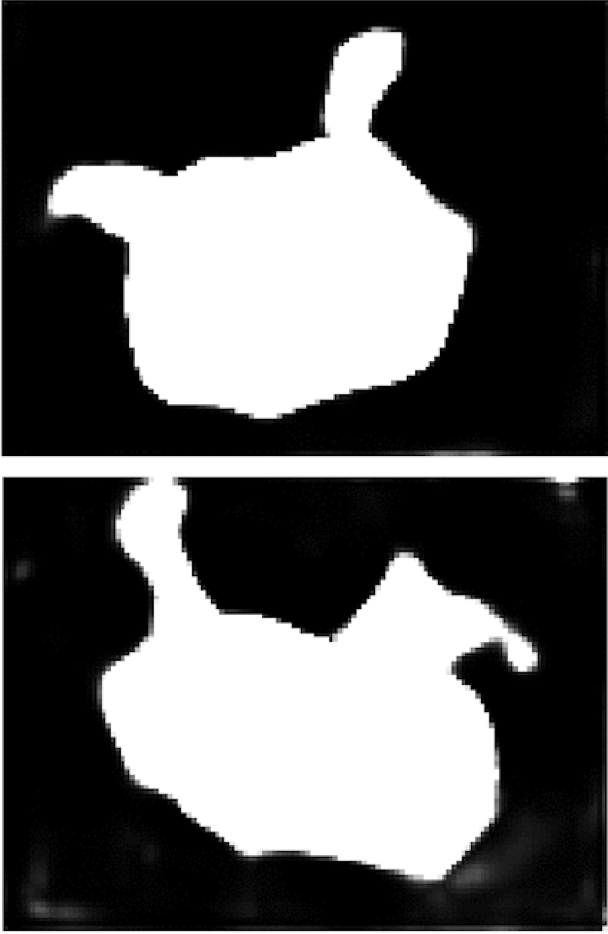}
        }
        \hspace*{-1.1em}
        \subfigure[]{
        \includegraphics[width=.17\textwidth]{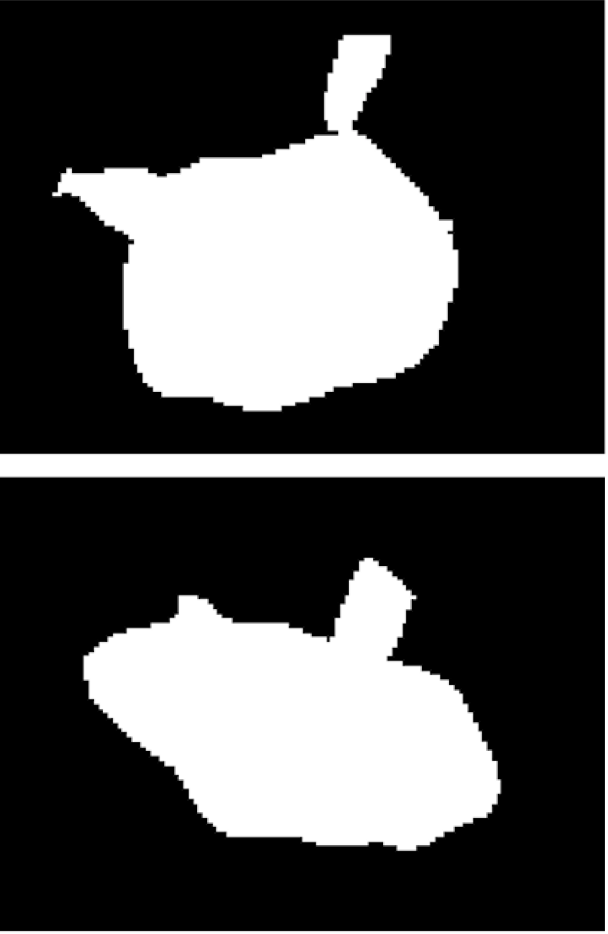}
        }
        \caption{Sparse volume segmentation and completion results for 2 cases. (a) Sparse ICE volume; (b) Completed CT volume; (c) the paired ``ground truth'' CT volume; (d) Predicted and (e) Ground truth 3D segmentation map.}
        \label{fig:3d_results}
    \end{figure}

\noindent \textbf{Dataset and preprocessing} The left atrial ICE images used in this study are collected using a clinical system with each image associated with a homogeneous matrix that projects the ICE image to a common coordinate system. We perform both 2D and 3D annotations on the ICE images for the cardiac components of interest, i.e., LA, LAA, LIPV, LSPV, RIPV and RSPV. For the 2D annotations, contours of all the plausible components in the current view are annotated. For the 3D annotations, ICE images, from the same patient and at the same cardiac phase \footnote{While in clinical practice multiple 2D ICE clips are acquired to dynamically image a patient's LA anatomy, here we focus on a stack of 2D ICE images, with often one gated frame per clip, and leave dynamic modeling for future study.}, are first projected to 3D, and 3D mesh models of the target components are then manually annotated. 3D segmentation masks are generated using these mesh models. In total, the whole database has 150 patients. For each patient, there are 20-80 gated frames for use. We have 3D annotations for all 150 patients. For 2D annotations, we annotated 100 patients, resulting in a total of 11,782 annotated ICE images. By anatomical components, we have in 2D 4669 LA, 1104 LAA, 1799 LIPV, 1603 LSPV, 1309 RIPV, and 1298 RSPV annotations. So, the LA is mostly observed and the LAA and PVs are less observed. For a subset of 1568 2D ICE images, we have 2-3 expert annotations per image to compute the inter-rater reliability (IRR).

As we do not have dense ICE volumes available for training, we use CT volumes instead as the ground truth for the completion task. Each CT volume is associated with a LA mesh model. To pair with a sparse ICE volume, we pick the CT volume whose LA mesh model is closest to that of the targeting sparse ICE volume (after Procrustes analysis \cite{schonemann1966generalized}). In total, 414 CT volumes are available, which gives enough anatomical variability for the mesh pairing. All the data used for 3D training are augmented with random perturbations in scale, rotation and translation to increase the generalizability of the model. \vspace{0.5em}

\noindent \textbf{Training and evaluation} We train the 3D-SCNet and 2D-RefineNet using Adam optimization with $lr = 0.005$, $\beta_1 = 0.5$, $\beta_2 = 0.999$. The 3D-SCNet is trained for about $25$ epochs with $\lambda^{s}_{adv} = 0.2$, $\lambda^{c}_{adv} = 1$, $\lambda^{s}_{rec} = 1000$, $\lambda^{c}_{rec} = 100$. The 2D-RefineNet is also trained for about $25$ epochs with $\lambda^{r}_{adv} = 1$,  $\lambda^{r}_{rec} = 1000$. All $\lambda$s are chosen empirically and we train the models using 5-fold cross-validation. The segmentation results are evaluated using the Dice metric and average symmetric surface distance (ASSD). \vspace{0.5em}

\begin{figure}[t]
    \centering
    \subfigure[Ground truth]{
    \includegraphics[width=.216\textwidth]{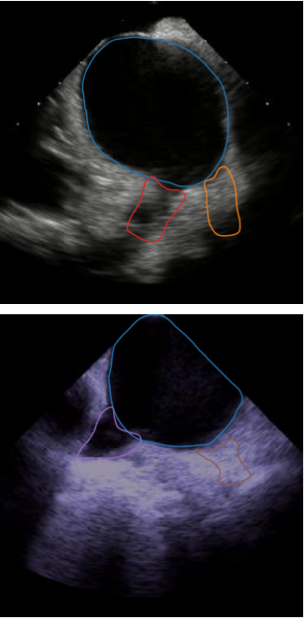}
    }
    \hspace*{-1.1em}
    \subfigure[2D only]{
    \includegraphics[width=.2174\textwidth]{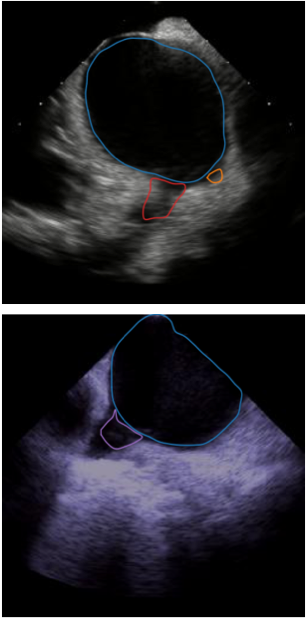}
    }
    \hspace*{-1.1em}
    \subfigure[3D only]{
    \includegraphics[width=.216\textwidth]{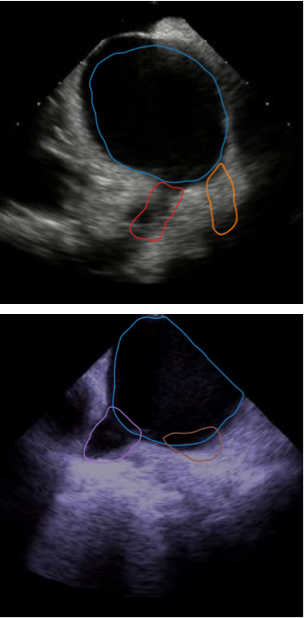}
    }
    \hspace*{-1.1em}
    \subfigure[2D + 3D]{
    \includegraphics[width=.216\textwidth]{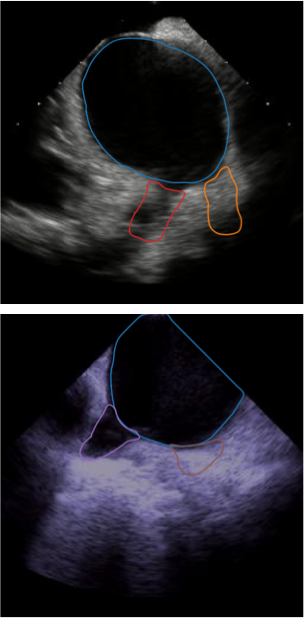}
    }
    \caption{Samples of 2D ICE contouring results from different models.}
    \label{fig:2d_results}
\end{figure}

\noindent \textbf{Results} The outputs from the 3D network model are shown in Fig. \ref{fig:3d_results}. We can observe that the model not only gives satisfying segmentation outputs, Fig. \ref{fig:3d_results}(d), but also gives a good estimation about the CT volume, Fig. \ref{fig:3d_results}(b). Especially, we note that the estimated completion outputs do not give structurally exact results as the ``ground truth'' but instead try to match the content from the sparse volume. Since the ``ground truth'' CT volume is paired based on mesh models, this difference is expected. It demonstrates that the completion outputs are based on the sparse volume and the system only tries to complete the missing region such that it looks like a ``real'' CT volume. We also quantitatively evaluate the performance of the 3D sparse volume segmentation and obtain the following Dice scores: LA (89.5\%), LAA (50.0\%), LIPV (52.9\%), LSPV (43.4\%), RIPV (62.43\%), RSPV (57.6\%) and overall (86.1\%). This shows that using the limited information from sparse volumes our model still can achieve a satisfactory 3D segmentation performance.  As we will show in later experiments, the segmentation accuracy, actually, is even higher in the region where 2D ICE images are presented. We also notice that \textit{it is vital to use the 3D appearance information} -- the training fails to converge in our experiment of learning the 3D network without using the 3D appearance information from CT.

Fig. \ref{fig:2d_results} shows the 2D ICE contouring results using different models: the ``2D only''  model that is trained directly with the 2D ICE images, the ``3D only''  model by projecting the predicted 3D segmentation results onto the corresponding ICE image, and the ``2D + 3D'' model by refining the outputs from 3D-SCNet using 2D-RefineNet. We observe from the first row that the ``3D only'' outputs give better estimation about the PVs (red and orange) than the ``2D only'' outputs. This is because the PVs in the current 2D ICE view are not clearly presented which is challenging for the ``2D only'' model. While for the ``3D only'' model, it makes use of the information from other views and hence predicts better the PV locations. Finally, we see that the outputs from the ``2D + 3D'' model combines the knowledge from both the 2D and 3D models and generally gives superior outputs than these two models. Similar results can also be found in the second row where we see the ``2D + 3D'' model not only predicts the location of the PVs (purple and brown) better by making use of the 3D information but also refines the output according to the 2D view.

\begin{table}[t]
    \centering
    \caption{2D segmentation accuracy of different models. The results are evaluated in terms of Dice metric (\%) and ASSD (mm).}
    \vspace{1em}
\begin{tabular}{@{}llclclclclclclclc@{}}
\toprule
      &  & \multicolumn{3}{c}{2D only} &  & \multicolumn{3}{c}{3D only}   &  & \multicolumn{3}{c}{2D+3D}         &  & \multicolumn{3}{c}{IRR} \\ \cmidrule(lr){3-5} \cmidrule(lr){7-9} \cmidrule(lr){11-13} \cmidrule(l){15-17} 
      &  & Dice    &    & ASSD    &  & Dice &  & ASSD           &  & Dice          &  & ASSD           &  & Dice    &     & ASSD    \\ \midrule
LA    &  & 94.3    &    & 0.623   &  & 93.5 &  & 0.693          &  & \textbf{95.4} &  & \textbf{0.537} &  & 89.6    &     & 1.340   \\
LAA   &  & 68.2    &    & 1.172   &  & 66.5 &  & 1.206          &  & \textbf{71.2} &  & \textbf{1.106} &  & 68.8    &     & 1.786   \\
LIPV  &  & 70.1    &    & 0.918   &  & 71.7 &  & 0.904          &  & \textbf{72.4} &  & \textbf{0.856} &  & 69.9    &     & 1.459   \\
LSPV  &  & 65.9    &    & 1.275   &  & 67.8 &  & \textbf{0.916} &  & \textbf{71.1} &  & 1.197          &  & 62.9    &     & 1.582   \\
RIPV  &  & 69.6    &    & 0.927   &  & 71.7 &  & 0.889          &  & \textbf{73.8} &  & \textbf{0.786} &  & 71.4    &     & 1.378   \\
RSPV  &  & 63.3    &    & 0.872   &  & 70.4 &  & \textbf{0.824} &  & \textbf{70.5} &  & 0.862          &  & 57.8    &     & 1.633   \\ \midrule
Total &  & 91.0    &    & 0.839   &  & 89.8 &  & 0.834          &  & \textbf{92.1}          &  & \textbf{0.791}          &  & 88.6    &     & 1.432   \\ \bottomrule
\end{tabular}
\label{tab:segmentation}
\vspace{-0.2in}
\end{table}

The quantitative results of these models are given in Table \ref{tab:segmentation}. The ``3D only'' model in general has better performance in PVs and worse performance in LA and LAA than the ``2D only'' model. This is because LA and LAA usually have a clear view in 2D ICE images, unlike the PVs. The ``2D + 3D'' model combines the advantages of the ``2D only'' and ``3D only'' model and in general yields the best performance. The IRR scores from human experts are relatively lower, especially for the LSPV and RSPV. This is expected as these two structures are difficult to view with ICE. The IRR scores are generally lower than those from our models, which demonstrates the benefit of using an automatic segmentation model -- better consistency.



\section{Conclusions and Future Work}

We present a knowledge fusion + deep learning approach to ICE contouring of multiple LA components. It uses 3D geometry and cross-modality appearance knowledge for better anatomical understanding and structural consistency. Then, it refines the contours in 2D by exploiting the detailed 2D appearance information. We show that the proposed model indeed benefits from the integrated knowledge and gives superior performance to the models trained individually. In the future, we will investigate the use of temporal information for better modeling and the clinical utility of the generated dense 3D cross-modality views. \newline

    
\bibliographystyle{splncs03}
\bibliography{references}

\begin{thebibliography}{10}
\providecommand{\url}[1]{\texttt{#1}}
\providecommand{\urlprefix}{URL }

\bibitem{allan2017simultaneous}
Allan, G., Nouranian, S., Tsang, T., Seitel, A., Mirian, M., Jue, J., Hawley,
  D., Fleming, S., Gin, K., Swift, J., et~al.: Simultaneous analysis of {2D}
  echo views for left atrial segmentation and disease detection. IEEE TMI
  36(1),  40--50 (2017)

\bibitem{bartel2013intracardiac}
Bartel, T., M{\"u}ller, S., Biviano, A., Hahn, R.T.: Why is intracardiac
  echocardiography helpful? {B}enefits, costs, and how to learn. European Heart
  Journal  35(2),  69--76 (2013)

\bibitem{caruana1998multitask}
Caruana, R.: Multitask learning. In: Learning to learn, pp. 95--133. Springer
  (1998)

\bibitem{funka2019three}
Funka-Lea, G., Liao, H., Zhou, S.K., Zheng, Y., Tang, Y.: Three-dimensional
  segmentation from two-dimensional intracardiac echocardiography imaging
  (Aug~29 2019), uS Patent App. 16/130,320

\bibitem{isola2017image}
Isola, P., Zhu, J.Y., Zhou, T., Efros, A.A.: Image-to-image translation with
  conditional adversarial networks. In: Proc. CVPR. pp. 1125--1134 (2017)

\bibitem{liao2018face}
Liao, H., Funka-Lea, G., Zheng, Y., Luo, J., Zhou, S.K.: Face completion with
  semantic knowledge and collaborative adversarial learning. In: Asian
  Conference on Computer Vision. pp. 382--397. Springer (2018)

\bibitem{lin2003combinative}
Lin, N., Yu, W., Duncan, J.S.: Combinative multi-scale level set framework for
  echo image segmentation. Medical Image Analysis  7(4),  529--537 (2003)

\bibitem{pathak2016context}
Pathak, D., Krahenbuhl, P., Donahue, J., Darrell, T., Efros, A.A.: Context
  encoders: Feature learning by inpainting. In: Proc. CVPR. pp. 2536--2544
  (2016)

\bibitem{ronneberger2015u}
Ronneberger, O., Fischer, P., Brox, T.: U-net: Convolutional networks for
  biomedical image segmentation. In: Proc. MICCAI. pp. 234--241. Springer
  (2015)

\bibitem{sanchez2014left}
S{\'a}nchez-Quintana, D., L{\'o}pez-M{\'\i}nguez, J.R., Mac{\'\i}as, Y.,
  Cabrera, J.A., Saremi, F.: Left atrial anatomy relevant to catheter ablation.
  Cardiology Research and Practice  (2014)

\bibitem{sarti2005maximum}
Sarti, A., Corsi, C., Mazzini, E., Lamberti, C.: Maximum likelihood
  segmentation of ultrasound images with rayleigh distribution. IEEE
  Transactions on Ultrasonics, Ferroelectrics, and Frequency Control  52(6),
  947--960 (2005)

\bibitem{schonemann1966generalized}
Sch{\"o}nemann, P.H.: A generalized solution of the orthogonal procrustes
  problem. Psychometrika  31(1),  1--10 (1966)

\bibitem{zhou2010shape}
Zhou, S.K.: Shape regression machine and efficient segmentation of left
  ventricle endocardium from {2D B}-mode echocardiogram. Medical Image Analysis
   14(4),  563--581 (2010)

\bibitem{zhou17ap}
Zhou, S., Shen, D., Greenspan, H. (eds.): Deep learning for medical image
  analysis. Academic Press (2017)

\bibitem{zoni2014}
Zoni-Berisso, M., Lercari, F., Carazza, T., Domenicucci, S.: Epidemiology of
  atrial fibrillation: European perspective. Clinical Epidemiology  6,
  213–20 (2014)

\end{thebibliography}

\end{document}